\pdfoutput=1

\def\thetitle{An Ensemble Method to Produce High-Quality Word Embeddings}

\def\scoreRW{.596}   

\documentclass[11pt,letterpaper]{article}
\usepackage{naaclhlt2016}
\usepackage{times}
\usepackage{latexsym}
\usepackage{amsmath}
\usepackage{booktabs}

\naaclfinalcopy 
\def\naaclpaperid{***} 

\usepackage{CJKutf8}
\usepackage{url}
\usepackage{graphicx}
\title{\thetitle}
\author{Robyn Speer\\
    Luminoso Technologies, Inc.\\
    675 Massachusetts Ave.\\
    Cambridge, MA 02139\\
    \texttt{rspeer@luminoso.com}
\And
    Joshua Chin\\
    Union College\\
    807 Union St.\\
    Schenectady, NY 12308\\
    \texttt{joshuarchin@gmail.com}
}

\date{April 6, 2016}

\begin{document}

\maketitle
\begin{abstract}

A currently successful approach to computational semantics is to represent
words as embeddings in a machine-learned vector space. We
present an ensemble method that combines embeddings produced by GloVe
\cite{pennington2014glove} and word2vec \cite{mikolov2013word2vec} with
structured knowledge from the semantic networks ConceptNet
\cite{speer2012conceptnet} and PPDB \cite{ganitkevitch2013ppdb}, merging their
information into a common representation with a large, multilingual vocabulary.
The embeddings it produces achieve state-of-the-art performance on many word-similarity evaluations. Its score of $\rho = \scoreRW{}$ on an evaluation of
rare words \cite{luong2013rw} is 16\% higher than the previous best known
system.

\end{abstract}

\section{Introduction}

Vector space models are an effective way to express the meanings of
natural-language terms in a computational system. These models are created
using machine-learning techniques that represent words or phrases as
vectors in a high-dimensional space, such that the cosine similarity of any two
terms corresponds to their semantic similarity.

These vectors, referred to as
the {\em embeddings} of the terms in the vector space, can also be used as an input to further steps of
machine learning. When algorithms expect dense vectors as input, embeddings
provide a representation that is both more compact and more informative than the
``one-hot'' representation in which every term in the vocabulary gets its own
dimension.

This kind of vector space has been used in applications such as search, topic
detection, and text classification, dating back to the introduction of latent
semantic analysis \cite{deerwester1990indexing}.  In recent years, there has
been a surge of interest in natural-language embeddings, as machine-learning
techniques such as \newcite{mikolov2013word2vec}'s word2vec and
\newcite{pennington2014glove}'s GloVe have begun to show dramatic improvements.
Word embeddings are often suggested as an initialization for more complex
methods, such as the sentence encodings of \newcite{kiros2015skip}.

\newcite{faruqui2015retrofitting} introduced a technique
known as ``retrofitting'', which combines embeddings learned from the
distributional semantics of unstructured text with a source of structured
connections between words. The combined embedding achieves performance on
word-similarity evaluations superior to either source individually.

Here, we build on the retrofitting process to produce a high-quality space of
word embeddings. We extend existing techniques in the following ways:

\begin{itemize}
\item We modify the retrofitting algorithm, making it not depend on the row
order of its input matrix, and allowing it to propagate over the union of the
vocabularies. This allows retrofitting to benefit from structured links outside
the original vocabulary, such as translations into other languages.
We call this procedure ``expanded retrofitting''.
\item We include ConceptNet, a Linked Open Data semantic network that expresses
many kinds of relationships between words in many languages, as a source of
structured connections between words.
\item We align English terms from different sources using a lemmatizer and a
heuristic for merging together multiple term vectors.
\item We fill gaps when aligning the two distributional-semantics sources
(GloVe and word2vec) using a locally linear interpolation.
\item We re-scale the distributional-semantics features using $L_1$ normalization.
\end{itemize}

When we use this process to combine word2vec, GloVe, PPDB, and ConceptNet, this
process produces a space of multilingual term embeddings we call the
``ConceptNet vector ensemble'' that achieves state-of-the-art performance on
word-similarity evaluations\footnote{Some methods and evaluations
\cite{agirre2009similarity} distinguish word similarity from word {\em
relatedness}. ``Coffee'' and ``mug'', for example, are quite related, but
not actually similar because coffee is not {\em like} a mug. In this paper,
however, we conflate similarity and relatedness into the same metric, as most
evaluations do.} over both common and rare words.

\subsection{Related Work}


\newcite{agirre2009similarity} observes that distributional similarity and
structured knowledge can be combined for a benefit exceeding what each would
achieve alone, particularly by extending the vocabulary. Their system uses a
similarity measure over WordNet, and uses distributional similarities to
recognize words outside of WordNet's vocabulary.

\newcite{levy2015embeddings} surveys modern methods of distributional
similarity and experiments with training them on specific data while varying
their parameters. They compare word2vec and GloVe, tune their hyperparameters
in a way that particularly improves word2vec, then proposes a method based on
the SVD of the Pointwise Mutual Information matrix that outperforms both. We
use Levy's results as a point of comparison here.

AutoExtend \cite{rothe2015autoextend} is a system with similar methods to
ours: it extends word2vec embeddings to cover all the word senses and synsets of
WordNet by propagating information over edges, thus combining distributional and
structured data after the fact. The primary goal of AutoExtend is word sense
disambiguation, and as such it is optimized for and evaluated on WSD tasks.
Our ensemble aims to extend and improve a vocabulary of undisambiguated words,
so there is no direct comparison between AutoExtend's results and ours.

\section{Knowledge Sources}

\subsection{ConceptNet and PPDB}
ConceptNet \cite{speer2012conceptnet} is a semantic network of terms
connected by labeled relations. Its terms are words or multiple-word phrases
in a variety of natural languages. For continuity with previous work,
these terms are often referred to as {\em concepts}.

ConceptNet originated as a machine-parsed version of the early crowd-sourcing
project called Open Mind Common Sense (OMCS) \cite{singh2002omcs}, and has expanded
to include several other data sources, both crowd-sourced and expert-created,
by unifying their vocabularies into a single representation. ConceptNet now includes
representations of WordNet \cite{miller1998wordnet}, Wiktionary \cite{wiktionary2014en},
and JMDict \cite{breen2004jmdict}, as well as data from ``games with a purpose'' in
multiple languages \cite{vonahn2006verbosity,kuo2009petgame,nakahara2011nadya}.
We choose not to include ConceptNet's alignment to DBPedia
\cite{auer2007dbpedia}
here, as DBPedia focuses on relations between specific named entities, which do not help
with general word similarity.\footnote{
    Given different goals -- such as achieving a high score on
    \newcite{mikolov2013word2vec}'s analogy evaluation that tests for
    implicit relations such as ``{\em A} is the CEO of company {\em B}'' --
    including an appropriate representation of DBPedia would of course be helpful.
}

PPDB \cite{ganitkevitch2013ppdb} is another resource that is useful for
learning about word similarity, providing different information from
ConceptNet. It lists pairs of words that are translated to the same word in parallel
corpora, particularly in documents of the European Parliament. PPDB is used
as an external knowledge source by \newcite{faruqui2015retrofitting}, so
we have evaluated the effect of adding it to our ensemble as well. As it seems to have
a small beneficial effect, we include it as part of the full ensemble.

\subsection{word2vec and GloVe}

word2vec and GloVe are two current systems that learn vector representations
of words according to their distributional semantics. Given a large text corpus,
they produce vectors representing similarities in how the words co-occur with
other words.

\newcite{mikolov2013word2vec} described a system of distributional word
embeddings called Skip-Grams with Negative Sampling (SGNS), which is more
popularly known by the name of its software implementation, {\em word2vec}.
(The {\em word2vec} software also implements another representation, Continuous
Bag-of-Words or CBOW, which is less often used for word similarity.)

In SGNS, a neural network with one hidden layer is trained to recognize words
that are likely to appear near each other. Its goal is to output a high value
when given examples of co-occurrences that appear in the data, and a low value
for negative examples where one word is replaced by a random word. The loss
function is weighted by the frequencies of the words involved and the distance
between them in the data. The word2vec
software\footnote{\url{https://code.google.com/p/word2vec/}} comes with SGNS
embeddings of text from Google News.

GloVe \cite{pennington2014glove} is an unsupervised learning algorithm that
learns a set of word embeddings such that the dot product of two words'
embeddings is approximately equal to the logarithm of their co-occurrence count.
The algorithm operates on a global word-word co-occurrence matrix, and
solves an optimization problem to learn a vector for each word, a separate
vector for each context (although the contexts are also words), and a bias
value for each word and each context. Only the word vectors are used for
computing similarity.

The embeddings that GloVe learns from data sources such as the Common
Crawl\footnote{\url{http://commoncrawl.org/}} are distributed on the GloVe web
page\footnote{\url{http://nlp.stanford.edu/projects/glove/}}. Here we evaluate
two downloadable sets of GloVe 1.2 embeddings, built from 42 billion and 840
billion tokens of the Common Crawl, respectively.

There is some debate about whether GloVe or word2vec is better at representing
word meanings in general. GloVe is presented by \newcite{pennington2014glove}
as performing better than word2vec on word-similarity tasks, but
\newcite{levy2015embeddings} finds that word2vec performs better with an
optimized setting of hyperparameters than GloVe does, when retrained with a
particular corpus.

In this paper, we focus only on the downloadable sets of term embeddings that
the GloVe and word2vec projects provide, not on re-running them with tuned
hyperparameters. Using this data makes it possible to reproduce their results
and compare directly to them, even when their preferred input data is not
available. We find that we can get very good results derived from
the downloadable embeddings, and that GloVe's downloadable embeddings outperform
word2vec's in this case, but a combination of them can perform even better.

\section{Methods}

\begin{figure}
\centering
\includegraphics[width=3.0in]{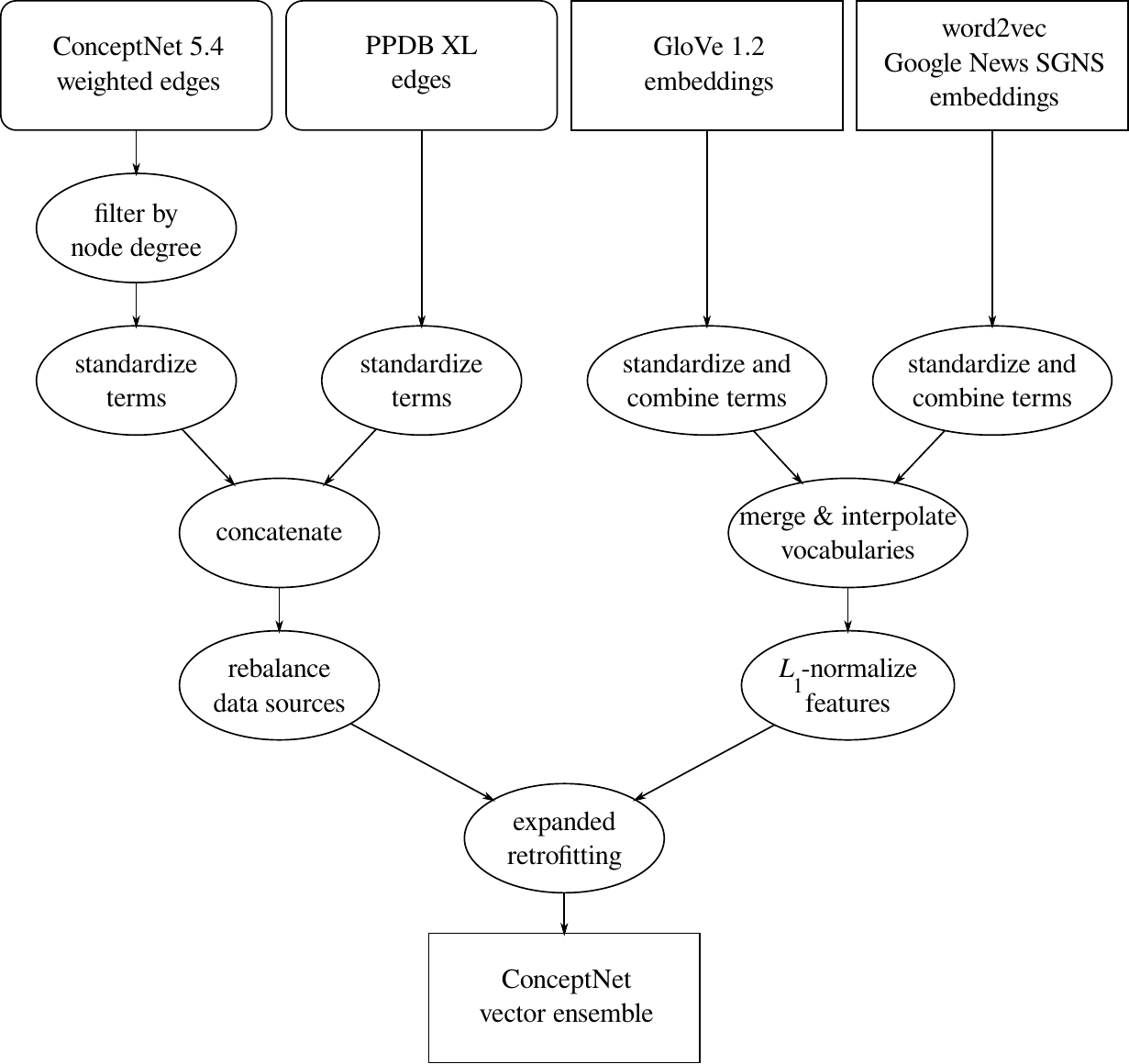}
\caption{
    The flow of data in building the ConceptNet vector ensemble from its
    data sources.
}
\label{dataflow}
\end{figure}

\newcite{faruqui2015retrofitting} introduced the ``retrofitting'' procedure,
which adjusts dense matrices of embeddings (such as the GloVe output) to take
into account external knowledge from a sparse semantic network. They tried
various sources of external knowledge, and the one that was most helpful to
GloVe was PPDB.  We found using ConceptNet to be more effective, and that
further marginal improvements could be achieved on some evaluations by
combining ConceptNet and PPDB.

Our goal is to create a 300-dimensional vector space that represents terms based
on a combination of GloVe and word2vec's downloadable embeddings, and structured
data from ConceptNet and PPDB. The resulting vector space allows information to
be shared among these various representations, including words that were not in
the vocabulary of the original representations. This includes low-frequency words
and even words that are not in English.

The complete process of building this vector space, whose steps will be explained
throughout this paper, appears in Figure~\ref{dataflow}.

As \newcite{levy2015embeddings} notes,
``[\ldots] much of the performance gains of word embeddings are due to certain
system design choices and hyperparameter optimizations, rather than the
embedding algorithms themselves.'' While it is presented as a negative result,
this simply emphasizes the importance of these system design choices.

Indeed, we have found that choices about how to handle terms and their
embeddings have a significant impact on evaluation results. One of these choices
involves how to pre-process words and phrases before looking them up, and
another involves the scale of the various features in the embeddings.

\subsection{Transforming and Aligning Vocabularies}
\label{standardizing-text}

\begin{figure}
\centering
\includegraphics[width=3.0in]{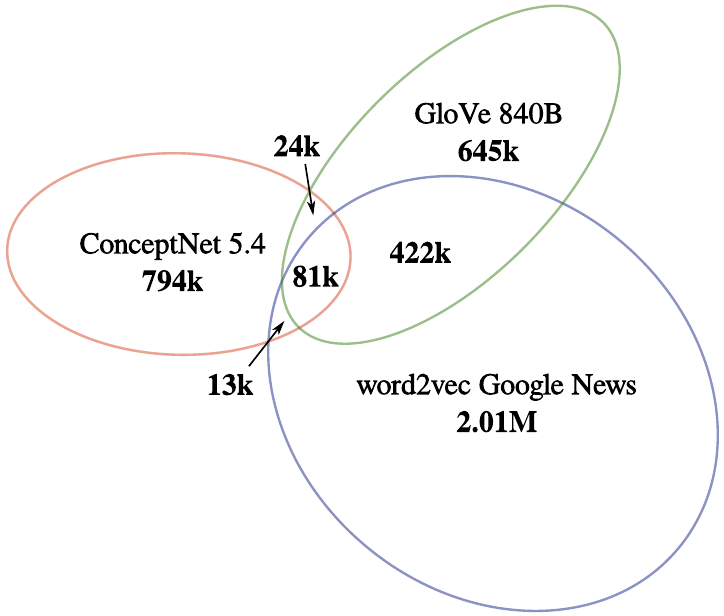}
\caption{
    A proportional-area diagram showing the overlap of vocabularies among
    ConceptNet and the available embeddings for word2vec and GloVe.
}
\label{vocabulary-overlap}
\end{figure}

Different representations apply
different pre-processing steps, placing strings in different equivalence
classes. We can only properly combine these resources if these string
representations are comparable to each other.

Pre-processing steps that various resources apply include: tokenizing text to
separate words from punctuation (which all inputs except GloVe 840B do),
joining multi-word phrases with underscores (ConceptNet and word2vec), removing
a small list of stopwords from multi-word phrases (ConceptNet only), folding the text to lowercase
(ConceptNet and GloVe 42B), replacing multiple digits with the character {\tt
\#} (word2vec only), and lemmatizing English words to their root form using a
modification of WordNet's Morphy algorithm (ConceptNet only).

We adapt a text pre-processing function from ConceptNet to apply a combination
of all of these processes, yielding a set of standardized, language-tagged
labels. As an example, the text ``Giving an example'' becomes the standardized
form {\tt /c/en/give\_example}. Applying this combined pre-processing function
to all labels increases the alignment of the various resources while reducing
the size of the combined vocabulary.

Because the transformations are many-to-one, this has the effect that a single
transformed term can become associated with multiple embeddings in a single
vector space. We considered a few options for dealing with these merged terms,
such as keeping only the highest-frequency term, averaging the vectors together,
or taking a weighted average based on their word frequency.

We found in preliminary evaluations that the weighted average was the best
approach. The multiple rows contain valuable data that should not simply be
discarded, but lower-frequency rows tend to have lower-quality data.

When using pre-trained vectors, it is often the case that intermediate
computations that produced these vectors (such as word frequencies) are not
available. What we do instead is to infer approximate word frequencies from
the fact that both GloVe and word2vec output their vocabularies in descending
order of frequency. We approximate the frequency distribution by assuming that
the tokens are distributed according to Zipf's law \cite{zipf1949human}: the
$n$th token in rank order has a frequency proportional to $1/n$. We use these
proportions in the weighted average when combining multiple embeddings.

This process alone is a benefit on word-similarity evaluations, even without
combining any resources. For example, the Rare Words (RW) dataset
\cite{luong2013rw} tends to encounter terms that are poorly represented or
out-of-vocabulary in most word embeddings.  Lemmatizing them before looking
them up, and combining them with more frequently observed representations,
improves the evaluation results on these words, even though the process
loses the ability to distinguish some word forms. The raw GloVe 840B data gets
a Spearman correlation of $\rho = .146$ on the RW dataset, which increases to
$\rho = .494$ when its embeddings are standardized and combined in this way.

Figure~\ref{vocabulary-overlap} shows the size of the vocabularies of
ConceptNet, GloVe, and word2vec after this transformation, and the sizes of
the overlaps among them, using a proportional-area Venn diagram produced using
{\em eulerAPE} \cite{micallef2014euler}.

\subsection{Feature Normalization}

As briefly mentioned by \newcite{pennington2014glove}, $L_2$ normalization of
the columns (that is, the 300 features) of the GloVe matrix provides a notable
increase in performance. One effect of normalization is to increase the weight
of distinguishing features and reduce the impact of noisy features.  Features
are more distinguishing for the purpose of cosine similarity when they contain
a few large values and many small ones.

We find that $L_1$ normalization of GloVe performs even better than $L_2$
normalization. $L_1$ causes occasional large values to have a smaller impact on the norm
than $L_2$ normalization. When a learning method such as GloVe has provided
highly selective features, $L_1$ normalization allows us to use them more effectively
in measuring similarity.

\subsection{Retrofitting}

Retrofitting \cite{faruqui2015retrofitting} is a process of combining existing
word vectors with a semantic lexicon. While the original formulation expresses
the problem in terms of updates that propagate over a set of edges, we have
found it more convenient to express it and implement it in terms of an update
to a matrix.

The inputs to retrofitting are an initial $m \times n$ dense matrix of term
embeddings, $W^0$, and a list of known semantic relationships.

Faruqui et al.'s retrofitting procedure aims to minimize a sum of a word's
distance from its neighbors in the semantic network and its distance from its
original vector. Its implementation in code takes steps along the gradient
toward this minimum by iteratively updating one vector $w'_i$ at a time to be a
linear combination between its original position $w^0_i$ and the average of its
neighbors in the semantic network.

The advantage of this iterative update is that it only requires two copies of
$W$ in memory ($W^0$ and the current state) and converges quickly. A disadvantage
is that the results depend on the order in which the nodes of the graph are
iterated, which is arbitrary.

We instead choose to update the embeddings all at once by multiplying them by a
sparse matrix $S$ of semantic connections. Letting $m'$ be the size of the
merged vocabulary, $S$ is an $m' \times m'$ matrix containing positive weighted
values for terms that are known to be semantically related, and 0 otherwise.
The rows of $S$ are scaled to have a sum of 1. We then add 1 to its diagonal to
help new terms converge on a single vector, as described in more detail below.

Let $W^0$ be an $m' \times n$ matrix whose rows come from the original
embeddings if available, and are all zeroes for terms outside the vocabulary of
the original embeddings. $A$ is a diagonal matrix of weights in which $A_{ii}$
is 1 if term $i$ is in the original vocabulary, and 0 otherwise (allowing us to
keep terms near their original embeddings without also keeping out-of-vocabulary
terms near the zero vector). We can now update $W$
iteratively so that the next iteration of $W$ is a combination of its product
with $S$ and its weighted original state, followed by $L_2$ normalization of its
non-zero rows\footnote{
    We maintain $L_2$ normalization so that minimizing distance and maximizing
    cosine similarity are always linked.
}:
$$
W^{k+1} = \mathrm{normalize}\left[ \left( S W^k + A W^0 \right)\left( I + A \right)^{-1} \right]
$$

The diagonal of the $S$ matrix relates each term to itself. We have found that
adding 1 to the diagonal -- effectively adding ``self-loops'' to the semantic
network -- helps the expanded retrofitting process converge.  Without this
diagonal, terms that only appear in the semantic network, and not in the
original embedding space, would get their value only from their neighbors at
every step, because their ``original position'' is the zero vector. This causes
large oscillations that prevent convergence.  With the diagonal, each term
vector is influenced by the vector it had in the previous step.

As a practical effect of this, Section~\ref{variations} (Varying the System)
will show that expanded retrofitting with self-loops added to the diagonal
performs better on word-similarity evaluations than it does without, when
allowed to run for 10 steps of retrofitting.

\subsection{ConceptNet as an Association Matrix}

In order to apply the expanded retrofitting method, we need to consider the data in
ConceptNet as a sparse, symmetric matrix of associations between terms. What
ConceptNet provides is more complex than that, as it connects terms with a
variety of not-necessarily-symmetric, labeled relations.

\newcite{havasi2010color} introduced a vector space embedding of ConceptNet,
``spectral association'', that disregarded the relation labels for the purpose
of measuring the relatedness of terms. Previous embeddings of ConceptNet, such
as that of \newcite{speer2008analogyspace}, preserved the relations but were
suited mostly for direct similarity and inference, not for relatedness. Because
most evaluation data for word similarity is also evaluating relatedness, unless
there has been a specific effort to separate them \cite{agirre2009similarity},
we erase the labels as in spectral association.

Each assertion in ConceptNet corresponds to two entries in a sparse
association matrix $S$.  ConceptNet assigns a confidence score, or weight, to
each assertion.  These weights are not entirely comparable between the data
sources that comprise ConceptNet, so we re-scaled them so that the average
weight of each different data source is 1.

An assertion that relates term $i$ to term $j$ with adjusted weight $w$ will
contribute $w$ to the values of $S_{ij}$ and $S_{ji}$. If another assertion
relates the same terms with a different relation, it will add to that value.
This constructs a symmetric matrix $S$, but the matrix we actually use in
retrofitting is the asymmetric $S'$, whose rows have been $L_1$-normalized to
prevent high-frequency concepts from overwhelming the results.

Due to the structure of ConceptNet, there exists a large fringe of terms that are
poorly connected to other nodes. To make the sparse matrix and the size of the
overall vocabulary more manageable, we filter ConceptNet when building its
association matrix: we exclude all terms that appear fewer than 3 times, English
terms that appear fewer than 4 times, and terms with more than 3 words in them.

\subsection{Locally Linear Alignment}
\label{locally-linear-alignment}

In order to use both word2vec and GloVe at the same time, we need to align their
partially-overlapping vocabularies and merge their features. This is
straightforward to do on the terms that are shared between the two vocabularies,
but we would rather not lose the other terms, if we can later benefit from
learning more about those terms from ConceptNet.

Before merging features, we need to compute GloVe representations for terms
represented in word2vec but not GloVe, and vice versa. The way we do this
is inspired by \newcite{zhao2015learning}, who infer translations between
languages of unknown phrases using a locally-linear projection of known
translations of similar phrases. Instead of known translations, we have the
terms that overlap between word2vec and GloVe. Given a non-overlapping term,
we calculate its vector as the average of the vectors of the nearest
overlapping terms, weighted by their cosine similarity.

To combine the features of word2vec and GloVe, we first concatenate their
vectors into 600-dimensional vectors. We then discount redundancy between its
features by transforming these 600-dimensional vectors with a singular value
decomposition (SVD).  We factor the matrix $M$ of concatenated vectors as $M = U
\Sigma V^T$, then compute the new joint features as $U \Sigma^{1/2}$.

$U \Sigma$ would be an orthogonal rotation of the original features;
$U \Sigma^{1/2}$ reduces the effect of its largest eigenvalues, making
over-represented features relatively smaller.

As with many decisions we make in preparing this data, we evaluated the benefit
of this step on our development data sets.  Discounting redundancy by replacing
$\Sigma$ by $\Sigma^{1/2}$ provides a benefit on two out of three data sets for
evaluating word similarity, as shown in Section~\ref{variations}.

It is common to use SVD as a form of dimensionality reduction, by discarding
the smallest singular values and truncating the matrix accordingly.
Section~\ref{variations} shows that we can reduce the interpolated matrix to
from 600 dimensions to 450 or 300 dimensions without much loss in performance.

\section{Evaluation}

\subsection{Word-Similarity Datasets}

We evaluate our model's performance at identifying similar words using a
variety of word-similarity gold standards:

\begin{itemize}
\item MEN-3000 \cite{bruni2014men}, crowd-sourced similarity judgments for 3000
    word pairs.
\item The Stanford Rare Words (RW) dataset \cite{luong2013rw}, crowd-sourced
    similarity judgments for 2034 word pairs, with a bias toward uncommon words.
\item WordSim-353 \cite{finkelstein2001ws}, a widely-used corpus of similarity
    judgments for 353 word pairs.
\item RG-65 \cite{rubenstein1965rg}, a classic corpus of similarity judgments
    for 65 word pairs, which has additionally been translated into German
    \cite{gurevych2005german} and French \cite{joubarne2011french}.
\end{itemize}

In striving to maximize an evaluation metric, it is important to hold out some
data, to avoid overfitting to the data by modifying the algorithm and its
parameters. The metrics we focused on improving were our rank correlation with
MEN-3000, which emphasizes having high-quality representations of common words,
and RW, which emphasizes having a broad vocabulary.

MEN-3000 comes with a development/test split, where 1000 of the 3000 word pairs
are held out for testing. We applied a similar split to RW, setting aside a
sample of $1/3$ of its word pairs for testing. In particular, We set aside
every third row, starting from row 3, using the Unix command {\tt split -un
r/3/3 rw.txt}. Similarly, we split on {\tt r/1/3} and {\tt r/2/3} and
concatenated the results to get the remaining evaluation data.

We did not apply a development/test split to WordSim-353 or RG-65, as they are
already much smaller than MEN and RW.

For the resources where we applied a development/test split, we evaluated
decisions we made in the code -- such as those described in
Section~\ref{variations} -- using only the development set, to preserve the
integrity of the test set and avoid ``overfitting via code''. We then evaluated
the final ensemble, with various pieces enabled, all at once on the held-out
test data to produce the results in this paper.

\subsection{Results}

\begin{table}[t]
\centering
\footnotesize
\begin{tabular}{llllll|rrr}
\toprule
\multicolumn{6}{c|}{Ensemble components} & \multicolumn{3}{c}{Evaluations} \\
\bf CN&\bf PP&\bf St&\bf W& \bf G& $L_1$  & \bf RW  & \bf MEN & \bf  WS \\
\midrule
     &      &      &      & g    &        &    .448 &    .816 &    .759 \\  
     &      &      &      & g    & $L_1$  &    .457 &    .820 &    .766 \\  
     &      &      &      & G    &        &    .146 &    .787 &    .672 \\  
     &      &      &      & G    & $L_1$  &    .148 &    .789 &    .676 \\  
     &      &      & W    &      &        &    .371 &    .732 &    .624 \\  
     &      &      & W    &      & $L_1$  &    .374 &    .732 &    .622 \\  
\midrule
     &      & St   &      & g    &        &    .492 &    .815 &    .765 \\  
     &      & St   &      & g    & $L_1$  &    .513 &    .834 &    .794 \\  
     &      & St   &      & G    &        &    .494 &    .814 &    .763 \\  
     &      & St   &      & G    & $L_1$  &    .513 &    .840 &    .798 \\  
     &      & St   & W    &      &        &    .453 &    .778 &    .731 \\  
     &      & St   & W    &      & $L_1$  &    .452 &    .777 &    .732 \\  
     &      & St   & W    & G    & $L_1$  &    .525 &    .832 &    .778 \\  
\midrule
     & PP   & St   &      & G    & $L_1$  &    .561 &    .852 &    .806 \\  
     & PP   & St   & W    &      & $L_1$  &    .481 &    .800 &    .750 \\  
     & PP   & St   & W    & G    & $L_1$  &    .543 &    .847 &    .782 \\  
CN   &      & St   &      & G    & $L_1$  &    .581 &    .860 &\bf .818 \\  
CN   &      & St   & W    &      & $L_1$  &    .541 &    .813 &    .771 \\  
CN   &      & St   & W    & G    & $L_1$  &    .598 &\bf .862 &    .802 \\  
CN   & PP   & St   &      & G    & $L_1$  &    .584 &    .860 &\bf .818 \\  
CN   & PP   & St   & W    &      & $L_1$  &    .543 &    .812 &    .775 \\  
CN   & PP   & St   & W    & G    & $L_1$  &\bf .601 &    .861 &    .802 \\  
\bottomrule
\end{tabular}

\caption{
    Word-similarity results as various components of the ensemble are enabled.
    The results are the Spearman rank correlation ($\rho$) with the held-out
    test sets of RW and MEN-3000 and with WordSim-353.
    The components are: {\bf CN} = ConceptNet,
    {\bf PP} = PPDB, {\bf St} = standardized and lemmatized terms,
    {\bf W} = word2vec SGNS vectors, {\bf g} = GloVe 42B vectors,
    {\bf G} = GloVe 1.2 840B vectors, {\bf $L_1$} = $L_1$-normalized features.
}
\label{eval-bigtable}
\end{table}

\begin{table}[t]
\footnotesize
\centering
\begin{tabular}{l|rrr|rrr}
\toprule
                   & \multicolumn{3}{|c|}{Rare Words} & \multicolumn{3}{|c}{MEN-3000} \\
System             &       dev &     test &      all &      dev &     test &      all \\
\midrule
GloVe 42B          &      .489 &     .448 &     .477 &     .813 &     .816 &     .814 \\
Mod. GloVe         &      .536 &     .513 &     .528 &     .841 &     .840 &     .841 \\
\bf Full ensemble  & \bf  .593 & \bf .601 & \bf .596 &     .858 &     .861 &     .859 \\
Omit CN5           &      .533 &     .543 &     .536 &     .842 &     .847 &     .844 \\
Omit PPDB          &      .590 &     .598 &     .592 & \bf .858 & \bf .862 & \bf .860 \\
Omit GloVe         &      .545 &     .543 &     .545 &     .807 &     .812 &     .808 \\
Omit word2vec      &      .591 &     .583 &     .588 &     .857 &     .859 &     .858 \\
\bottomrule
\end{tabular}

\caption{
    A comparison of evaluation results between the ``dev'' datasets that were
    used in development, and the held-out ``test'' datasets, for selected systems.
}
\label{eval-dev-test}
\end{table}

Table~\ref{eval-bigtable} shows the performance of the ensemble as various
components of it are enabled. {\bf G} and {\bf g} indicate that the initial
embeddings come from GloVe (840B or 42B respectively), and {\bf W} indicates
that they are word2vec's SGNS embeddings built from Google News. When both
{\bf W} and {\bf G} are present, the embeddings are combined as in
Section~\ref{locally-linear-alignment}. $L_1$ indicates that the columns of
features were $L_1$-normalized; otherwise we used the existing scale of the
features.\footnote{
    When GloVe is not $L_1$-normalized, it is $L_2$-normalized instead,
    following \newcite{pennington2014glove}.
}

{\bf St} indicates that the labels were transformed and rows combined using the
method of Section~\ref{standardizing-text}, which is a prerequisite to combining
multiple data sources. {\bf CN} and {\bf PP} indicate adding data from
ConceptNet, PPDB, or both using expanded retrofitting.
Note that the row labeled with {\bf g} alone is simply an evaluation of GloVe
42B that reproduces the evaluation of \newcite{pennington2014glove}. Our results
here match the published results to within .001.

While Table~\ref{eval-bigtable} shows our correlation with only the test data on
these evaluations, Table~\ref{eval-dev-test} compares our results on
development and test data.

\subsection{Benefits of Lemmatization}

Decisions about how to process the data, even after
the fact, make a very large difference in word-similarity evaluations.
Comparing the GloVe 42B results to the 840B results (Table~\ref{eval-bigtable}), we see that GloVe 42B
works better ``out of the box'', and 840B contains messy data that particularly
causes problems on the Rare Words evaluation. However, our strategy to
standardize and lemmatize the term labels of GloVe 840B, combining its rows using
the Zipf estimate, makes it perform better than GloVe 42B and other published
results, as seen in the {\bf St}, {\bf G}, $L_1$ row. We call this
configuration ``Modified GloVe'',
and similarly, our best configuration of word2vec is ``Modified word2vec''.

The fact that Modified GloVe performs better than GloVe 42B, GloVe 840B, and many
other systems, even before retrofitting any additional data onto it, highlights
the unexpectedly large benefit of lemmatization: some of the improvements from
this paper can be realized without introducing any additional data.

It's important to note that we are not changing the evaluation data by using
a lemmatizer; we are only changing the way we look it up as embeddings in the
vector space that we are evaluating.
For example, if an evaluation requires
similarities for the words ``dry'' and ``dried'' to be ranked differently, or
the words ``polish'' and ``Polish'', the lemmatized system will rank them the
same, and will be penalized in its Spearman correlation.
However, the benefits of lemmatization when evaluating semantic similarity
appear to far outweigh the drawbacks.

\subsection{Comparisons to Other Published Results}

\begin{table}[t]
\centering
\footnotesize
\begin{tabular}{l|rrr}
\toprule
Method                        & RW [all]  &      MEN  & WS    \\
\midrule
word2vec SGNS (Levy)          &     .470  &     .774  & .733* \\
Modified word2vec (ours)      &     .476  &     .778  & .731  \\
GloVe (Pennington)            &     .477  &     .816  & .759  \\
Modified GloVe (ours)         &     .528  &     .840  & .798  \\
SVD (Levy)                    &     .514  &     .778  & .736* \\
Retrofitting (Faruqui)        &      ---  &     .796  & .741  \\
ConceptNet vector ensemble    & \bf .596  & \bf .859  & \bf .821 \\
\bottomrule
\end{tabular}

\caption{
    Comparison between our ensemble word embeddings and previous results,
    on the complete RW data, the MEN-3000 test data, and WordSim-353.
    Asterisks indicate estimated overall results for WordSim.
}
\label{compare-others}
\end{table}

\begin{figure}
\includegraphics[width=3.0in]{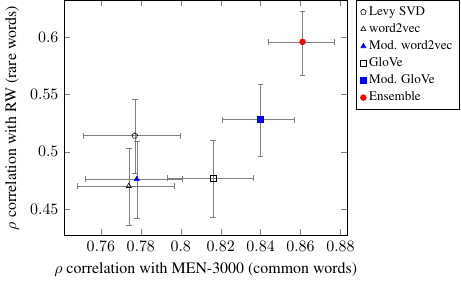}
\caption{
    Systems discussed in this paper, plotted according to their Spearman
    correlation with MEN-3000 and RW. Hollow nodes are previously-reported
    results; filled nodes are systems created for this paper.
}
\label{compare-graph}
\end{figure}

\begin{table}[t]
\centering
\footnotesize
\begin{tabular}{l|rrr}
\toprule
                                  & \multicolumn{3}{c}{RG-65 language} \\
Method                            &      en &      de &      fr \\
\midrule
Faruqui et al.: SG + Retrofitting &    .739 &    .603 &    .606 \\
ConceptNet vector ensemble        &\bf .891 &\bf .645 &\bf .789 \\
\bottomrule
\end{tabular}

\caption{
    Evaluation results comparing Faruqui et al.'s
    multilingual retrofitting of Wikipedia skip-grams to our ensemble,
    on RG-65 and its translations.
}
\label{eval-multilingual}
\end{table}

In Table~\ref{compare-others}, we compare our results on the RW and MEN-3000
datasets to the best published results that we know of.
\newcite{levy2015embeddings} present results including an SVD-based method that
scores $\rho = .514$ on the RW evaluation, as well as an implementation of
skip-grams with negative sampling (SGNS), originally introduced by
\newcite{mikolov2013word2vec}, with optimized hyperparameters. We also compare to the original results from
GloVe 42B, and the best MEN-3000 result from
\newcite{faruqui2015retrofitting}. We use the complete RW data, not our test
set, so that we can compare directly to previous results.

Levy's evaluation uses a version of WordSim-353 that is split into separate
sets for similarity and relatedness. We estimate the overall score using a
weighted average based on the size of the split datasets.

The RW and MEN data is also plotted in Figure~\ref{compare-graph}. Error bars
indicate 95\% confidence intervals based on the Fisher transformation of $\rho$
\cite{fisher1915frequency}, supposing that each evaluation is randomly sampled
from a hypothetical larger data set.

The ConceptNet vector ensemble with all six components performs better than the
previously published systems: we can reject the null hypothesis that the
ensemble performs the same as one of these published systems with $p < .01$.
It is inconclusive whether it is better to include or exclude PPDB in the
ensemble, as the results with and without it are very close.

Table~\ref{eval-multilingual} shows the performance of these systems on
gold standards that have been translated to other languages, in comparison to
the multilingual results published by \newcite{faruqui2015retrofitting}. Our
system performs well in non-English languages even though the vocabularies of
word2vec and GloVe are assumed to be English only. The representations of
non-English words come from expanded retrofitting, which allows information to
propagate over the inter-language links in ConceptNet.

\subsection{Varying the System}
\label{variations}

\begin{table*}[t]
\footnotesize
\centering
\begin{tabular}{lrrrrrr}
\toprule
Modification & RW [dev] & MEN [dev] & WS-353 \\
\midrule
{\bf Unmodified}                     & \bf .593 &     .858 &     .802 \\
Use first row instead of row-merging &     .572 &     .816 &     .764 \\
Unweighted row-merging               &     .582 &     .856 &     .779 \\
No self-loops in retrofitting        &     .563 &     .857 &     .764 \\
Interpolate without SVD              &     .579 &     .856 & \bf .813 \\
Reduce from 600 to 450 dimensions    &     .586 &     .858 &     .797 \\
Reduce from 600 to 300 dimensions    &     .583 & \bf .859 &     .791 \\
\bottomrule
\end{tabular}
\caption{
    The effects of various modifications to the full ensemble.
    RW and MEN-3000 were evaluated using their development sets here,
    not the held-out test data.
}
\label{eval-variations}
\end{table*}

Some of the procedures we implemented in creating the ensemble require some
justification. To show the benefits of certain decisions, such as adding
self-loops or the way we choose to merge rows of GloVe, we have evaluated what
happens to the system in the absence of each decision. These evaluations appear
in Table~\ref{eval-variations}. These evaluations were part of how we decided
on the best configuration of the system, so they were run on the development
sets of RW and MEN-3000, not the held-out test sets.

In Table~\ref{eval-variations}, we can see that the choice of how to merge rows
of GloVe that get the same standardized label makes a large difference.  Recall
that the method we ultimately used was to assign each row a pseudo-frequency
based on Zipf's law, and then take a weighted average based on those
frequencies.  The results drop noticeably when we try the other proposed
methods, which are taking only the first (most frequent) row that appears, or
taking the unweighted average of the rows.

The same table shows that we can save some computation time and space by
reducing the dimensionality of the feature vectors while building the matrix
that combines word2vec and GloVe. Reducing the dimensionality seems to cause a
small degradation in RW score, but the MEN and WordSim scores stay around the
same value, even increasing by an inconclusive amount in some cases.

If we skip applying the SVD at all in the interpolation step -- that is, when a
term has features in word2vec and GloVe, we simply concatenate those features
with all their redundancy -- it lowers the RW score somewhat, but raises the
WordSim-353 score.

In the retrofitting procedure, we made the decision to add ``self-loops'' that
connect each term to itself, because this helps stabilize the
representations of terms that are outside the original vocabulary of GloVe.
Reversing this decision (and running for the same number of steps) causes a
noticeable drop in performance on RW, the evaluation that is most likely to
involve words that were poorly represented or unrepresented in GloVe.

In separate experimentation, we found that when we separate ConceptNet into its
component datasets and drop each one in turn, the effects on the evaluation
results are mostly quite small. There is no single dataset that acts as the
``keystone'' without which the system falls apart, but one dataset ---
Wiktionary --- unsurprisingly has a larger effect than the others, because it
is responsible for most of the assertions. Of the 5,631,250 filtered assertions
that come from ConceptNet, 4,244,410 of them are credited to Wiktionary.

Without Wiktionary, the score on RW drops from .587 to .541. However, at the
same time, the MEN-3000 score {\em increases} from .858 to .865. The system
continues to do what it is designed to do without Wiktionary, but there seems to
be a tradeoff in performance on rare words and common words involved.

\section{Conclusions}

The work we have presented here involves building on many previous techniques,
while adding some new techniques and new sources of knowledge.

As \newcite{levy2015embeddings} found, high-level choices about how to use a
system can significantly affect its performance. While Levy found settings of
hyperparameters that made word2vec outperform GloVe, we found that we can make
GloVe outperform Levy's tuned word2vec by pre-processing the words in GloVe
with case-folding and lemmatization, and re-weighting the features using $L_1$
normalization.

We also showed that it isn't necessary to choose just one of word2vec or GloVe
as a starting point. Instead, we can benefit from both of them using a
locally-linear interpolation between them.

We showed that ConceptNet is a useful source of structured knowledge that was
not considered in previous work on {\em retrofitting} distributional semantics
with structured knowledge, especially when retrofitting is generalized into our
{\em expanded retrofitting} technique, which can benefit from links that are
outside the original vocabulary.

\subsection{Future Work}

One aspect of our method that clearly has room for improvement is the fact that
we disregard the labels on relations in ConceptNet.  There is valuable
knowledge there that we might be able to take into account with a more
sophisticated extension of retrofitting, one that goes beyond simply knowing
that particular words should be related, to handling them differently based on
{\em how} they are related, as in the RESCAL representation
\cite{nickel2011rescal}. This seems particularly important for antonyms, which
indicate that words are similar overall but different in a key aspect, such as
forming two ends of the same scale.

Lemmatization is clearly a useful component of our word-similarity
representation, but it loses information. Representing morphological
relationships as operations in the vector space, as in
\newcite{soricut2015unsupervised}, could yield a better representation of
similarities between forms of words.

We believe that the variety of data sources represented in ConceptNet helped to
improve evaluation scores by expanding the domain of the system's knowledge.
There's no reason the improvement needs to stop here. It is quite likely that
there are more sources of linked open data that could be included, or further
standardizations that could be applied to the text to align more data. An
appropriate representation of Universal WordNet \cite{demelo2009uwn} could
improve the multilingual performance, for example, as could embeddings built
from the distribution of words in non-English text. Adapting Rothe and
Schütze's AutoExtend representation could provide a representation of word
senses and a more specific understanding of words to the system.

\section{Reproducing These Results}

We aim for these results to be reproducible and reusable by others. The code
that we ran to produce the results in this paper is available in the GitHub
repository \url{https://github.com/LuminosoInsight/conceptnet-vector-ensemble}.
We have tagged this revision as {\em submitted-20160406}.

The README of that repository also points to URLs where the computed results
can be downloaded.

(As this paper is being updated in 2019, the exact data that went into building
the embeddings used in this paper has sadly been lost. However, the repository
points to newer data and newer code that implements an updated version of this
process.)

\section*{Acknowledgements}

We thank Catherine Havasi for overseeing and reviewing this work, and Avril Kenney,
Dennis Clark, and Alice Kaanta for providing feedback on drafts of this paper.

We thank the word2vec, GloVe, and PPDB teams for opening their data so that new
techniques can be built on top of their results. We also thank our
collaborators who have contributed code and data to ConceptNet over the years,
and the tens of thousands of pseudonymous contributors to Wiktionary, Open Mind
Common Sense, and related projects, for their frequently uncredited work in
providing freely-available lexical knowledge.

\begin{CJK*}{UTF8}{min}
\bibliography{wordsim_paper}

\begin{thebibliography}{}

\bibitem[\protect\citename{Agirre \bgroup et al.\egroup
  }2009]{agirre2009similarity}
Eneko Agirre, Enrique Alfonseca, Keith Hall, Jana Kravalova, Marius
  Pa{\c{s}}ca, and Aitor Soroa.
\newblock 2009.
\newblock A study on similarity and relatedness using distributional and
  {W}ord{N}et-based approaches.
\newblock In {\em Proceedings of Human Language Technologies: The 2009 Annual
  Conference of the North American Chapter of the Association for Computational
  Linguistics}, pages 19--27. Association for Computational Linguistics.

\bibitem[\protect\citename{Auer \bgroup et al.\egroup }2007]{auer2007dbpedia}
S{\"o}ren Auer, Christian Bizer, Georgi Kobilarov, Jens Lehmann, Richard
  Cyganiak, and Zachary Ives.
\newblock 2007.
\newblock {\em {DB}pedia: A nucleus for a web of open data}.
\newblock Springer.

\bibitem[\protect\citename{Breen}2004]{breen2004jmdict}
James Breen.
\newblock 2004.
\newblock {JM}{D}ict: a {J}apanese-multilingual dictionary.
\newblock In {\em Proceedings of the Workshop on Multilingual Linguistic
  Ressources}, pages 71--79. Association for Computational Linguistics.

\bibitem[\protect\citename{Bruni \bgroup et al.\egroup }2014]{bruni2014men}
Elia Bruni, Nam-Khanh Tran, and Marco Baroni.
\newblock 2014.
\newblock Multimodal distributional semantics.
\newblock {\em J. Artif. Intell. Res. (JAIR)}, 49:1--47.

\bibitem[\protect\citename{De~Melo and Weikum}2009]{demelo2009uwn}
Gerard De~Melo and Gerhard Weikum.
\newblock 2009.
\newblock Towards a universal wordnet by learning from combined evidence.
\newblock In {\em Proceedings of the 18th ACM conference on Information and
  knowledge management}, pages 513--522. ACM.

\bibitem[\protect\citename{Deerwester \bgroup et al.\egroup
  }1990]{deerwester1990indexing}
Scott~C. Deerwester, Susan~T Dumais, Thomas~K. Landauer, George~W. Furnas, and
  Richard~A. Harshman.
\newblock 1990.
\newblock Indexing by latent semantic analysis.
\newblock {\em JAsIs}, 41(6):391--407.

\bibitem[\protect\citename{Faruqui \bgroup et al.\egroup
  }2015]{faruqui2015retrofitting}
Manaal Faruqui, Jesse Dodge, Sujay~K. Jauhar, Chris Dyer, Eduard Hovy, and
  Noah~A. Smith.
\newblock 2015.
\newblock Retrofitting word vectors to semantic lexicons.
\newblock In {\em Proceedings of NAACL}.

\bibitem[\protect\citename{Finkelstein \bgroup et al.\egroup
  }2001]{finkelstein2001ws}
Lev Finkelstein, Evgeniy Gabrilovich, Yossi Matias, Ehud Rivlin, Zach Solan,
  Gadi Wolfman, and Eytan Ruppin.
\newblock 2001.
\newblock Placing search in context: The concept revisited.
\newblock In {\em Proceedings of the 10th international conference on World
  Wide Web}, pages 406--414. ACM.

\bibitem[\protect\citename{Fisher}1915]{fisher1915frequency}
Ronald~A Fisher.
\newblock 1915.
\newblock Frequency distribution of the values of the correlation coefficient
  in samples from an indefinitely large population.
\newblock {\em Biometrika}, pages 507--521.

\bibitem[\protect\citename{Ganitkevitch \bgroup et al.\egroup
  }2013]{ganitkevitch2013ppdb}
Juri Ganitkevitch, Benjamin Van~Durme, and Chris Callison-Burch.
\newblock 2013.
\newblock {PPDB}: The paraphrase database.
\newblock In {\em HLT-NAACL}, pages 758--764.

\bibitem[\protect\citename{Gurevych}2005]{gurevych2005german}
Iryna Gurevych.
\newblock 2005.
\newblock Using the structure of a conceptual network in computing semantic
  relatedness.
\newblock In {\em Natural Language Processing--IJCNLP 2005}, pages 767--778.
  Springer.

\bibitem[\protect\citename{Havasi \bgroup et al.\egroup }2010]{havasi2010color}
Catherine Havasi, Robyn Speer, and Justin Holmgren.
\newblock 2010.
\newblock Automated color selection using semantic knowledge.
\newblock In {\em AAAI Fall Symposium: Commonsense Knowledge}.

\bibitem[\protect\citename{Joubarne and Inkpen}2011]{joubarne2011french}
Colette Joubarne and Diana Inkpen.
\newblock 2011.
\newblock Comparison of semantic similarity for different languages using the
  {G}oogle {N}-gram corpus and second-order co-occurrence measures.
\newblock In {\em Advances in Artificial Intelligence}, pages 216--221.
  Springer.

\bibitem[\protect\citename{Kiros \bgroup et al.\egroup }2015]{kiros2015skip}
Ryan Kiros, Yukun Zhu, Ruslan~R Salakhutdinov, Richard Zemel, Raquel Urtasun,
  Antonio Torralba, and Sanja Fidler.
\newblock 2015.
\newblock Skip-thought vectors.
\newblock In {\em Advances in Neural Information Processing Systems}, pages
  3276--3284.

\bibitem[\protect\citename{Kuo \bgroup et al.\egroup }2009]{kuo2009petgame}
Yen-ling Kuo, Jong-Chuan Lee, Kai-yang Chiang, Rex Wang, Edward Shen, Cheng-wei
  Chan, and Jane Yung-jen Hsu.
\newblock 2009.
\newblock Community-based game design: experiments on social games for
  commonsense data collection.
\newblock In {\em Proceedings of the ACM SIGKDD Workshop on Human Computation},
  pages 15--22. ACM.

\bibitem[\protect\citename{Levy \bgroup et al.\egroup
  }2015]{levy2015embeddings}
Omer Levy, Yoav Goldberg, and Ido Dagan.
\newblock 2015.
\newblock Improving distributional similarity with lessons learned from word
  embeddings.
\newblock {\em Transactions of the Association for Computational Linguistics},
  3:211--225.

\bibitem[\protect\citename{Luong \bgroup et al.\egroup }2013]{luong2013rw}
Minh-Thang Luong, Richard Socher, and Christopher~D Manning.
\newblock 2013.
\newblock Better word representations with recursive neural networks for
  morphology.
\newblock {\em CoNLL-2013}, 104.

\bibitem[\protect\citename{Micallef and Rodgers}2014]{micallef2014euler}
Luana Micallef and Peter Rodgers.
\newblock 2014.
\newblock {eulerAPE}: Drawing area-proportional 3-venn diagrams using ellipses.
\newblock {\em PLoS ONE}, 9(7):e101717, 07.

\bibitem[\protect\citename{Mikolov \bgroup et al.\egroup
  }2013]{mikolov2013word2vec}
Tomas Mikolov, Kai Chen, Greg Corrado, and Jeffrey Dean.
\newblock 2013.
\newblock Efficient estimation of word representations in vector space.
\newblock {\em CoRR}, abs/1301.3781.

\bibitem[\protect\citename{Miller \bgroup et al.\egroup
  }1998]{miller1998wordnet}
George Miller, Christiane Fellbaum, Randee Tengi, P~Wakefield, H~Langone, and
  BR~Haskell.
\newblock 1998.
\newblock {\em WordNet}.
\newblock MIT Press Cambridge.

\bibitem[\protect\citename{Nakahara and Yamada}2011]{nakahara2011nadya}
Kazuhiro Nakahara and Shigeo Yamada.
\newblock 2011.
\newblock 日本でのコモンセンス知識獲得を目的とした{ W}eb
  ゲームの開発と評価 { }[{D}evelopment and evaluation of a {W}eb-based
  game for common-sense knowledge acquisition in {J}apan].
\newblock {\em Unisys 技報}, 30(4):295--305.

\bibitem[\protect\citename{Nickel \bgroup et al.\egroup
  }2011]{nickel2011rescal}
Maximilian Nickel, Volker Tresp, and Hans-Peter Kriegel.
\newblock 2011.
\newblock A three-way model for collective learning on multi-relational data.
\newblock In {\em Proceedings of the 28th international conference on machine
  learning (ICML-11)}, pages 809--816.

\bibitem[\protect\citename{Pennington \bgroup et al.\egroup
  }2014]{pennington2014glove}
Jeffrey Pennington, Richard Socher, and Christopher~D Manning.
\newblock 2014.
\newblock Glo{V}e: Global vectors for word representation.
\newblock {\em Proceedings of the Empiricial Methods in Natural Language
  Processing (EMNLP 2014)}, 12:1532--1543.

\bibitem[\protect\citename{Rothe and Sch\"{u}tze}2015]{rothe2015autoextend}
Sascha Rothe and Hinrich Sch\"{u}tze.
\newblock 2015.
\newblock Auto{E}xtend: Extending word embeddings to embeddings for synsets and
  lexemes.
\newblock In {\em Proceedings of the 53rd Annual Meeting of the Association for
  Computational Linguistics and the 7th International Joint Conference on
  Natural Language Processing (Volume 1: Long Papers)}, pages 1793--1803,
  Beijing, China, July. Association for Computational Linguistics.

\bibitem[\protect\citename{Rubenstein and Goodenough}1965]{rubenstein1965rg}
Herbert Rubenstein and John~B Goodenough.
\newblock 1965.
\newblock Contextual correlates of synonymy.
\newblock {\em Communications of the ACM}, 8(10):627--633.

\bibitem[\protect\citename{Singh \bgroup et al.\egroup }2002]{singh2002omcs}
Push Singh, Thomas Lin, Erik~T Mueller, Grace Lim, Travell Perkins, and Wan~Li
  Zhu.
\newblock 2002.
\newblock {O}pen {M}ind {C}ommon {S}ense: Knowledge acquisition from the
  general public.
\newblock In {\em On the move to meaningful internet systems 2002: CoopIS, DOA,
  and ODBASE}, pages 1223--1237. Springer.

\bibitem[\protect\citename{Soricut and Och}2015]{soricut2015unsupervised}
Radu Soricut and Franz Och.
\newblock 2015.
\newblock Unsupervised morphology induction using word embeddings.
\newblock In {\em Proc. NAACL}.

\bibitem[\protect\citename{Speer and Havasi}2012]{speer2012conceptnet}
Robyn Speer and Catherine Havasi.
\newblock 2012.
\newblock Representing general relational knowledge in {C}oncept{N}et 5.
\newblock In {\em LREC}, pages 3679--3686.

\bibitem[\protect\citename{Speer \bgroup et al.\egroup
  }2008]{speer2008analogyspace}
Robyn Speer, Catherine Havasi, and Henry Lieberman.
\newblock 2008.
\newblock Analogy{S}pace: Reducing the dimensionality of common sense
  knowledge.
\newblock In {\em AAAI}, volume~8, pages 548--553.

\bibitem[\protect\citename{von Ahn \bgroup et al.\egroup
  }2006]{vonahn2006verbosity}
Luis von Ahn, Mihir Kedia, and Manuel Blum.
\newblock 2006.
\newblock Verbosity: a game for collecting common-sense facts.
\newblock In {\em Proceedings of the SIGCHI conference on Human Factors in
  computing systems}, pages 75--78. ACM.

\bibitem[\protect\citename{Wiktionary}2014]{wiktionary2014en}
Wiktionary.
\newblock 2014.
\newblock Wiktionary{,} the free dictionary --- {E}nglish data export.
\newblock (A collaborative project with thousands of authors.) Retrieved from
  \url{https://dumps.wikimedia.org/enwiktionary/} on 2014-08-26.

\bibitem[\protect\citename{Zhao \bgroup et al.\egroup }2015]{zhao2015learning}
Kai Zhao, Hany Hassan, and Michael Auli.
\newblock 2015.
\newblock Learning translation models from monolingual continuous
  representations.
\newblock In {\em Proceedings of NAACL}.

\bibitem[\protect\citename{Zipf}1949]{zipf1949human}
G.K. Zipf.
\newblock 1949.
\newblock {\em Human behavior and the principle of least effort: an
  introduction to human ecology}.
\newblock Addison-Wesley Press.

\end{thebibliography}
\end{CJK*}

\end{document}